\documentclass{article}

\usepackage{subfig, hyperref, times}

\newcommand{\titlename}{Gaussian Mean Field Regularizes by Limiting Learned Information}

\title{\titlename}

\usepackage{tabularx,tabulary}

\usepackage{svg}

\usepackage{tikz}
\usetikzlibrary{bayesnet}

\usepackage{cleveref}

\usepackage{wrapfig}

\newcommand{\icmlheader}{\twocolumn[
\icmltitle{\titlename}

\icmlsetsymbol{equal}{*}

\begin{icmlauthorlist}
\icmlauthor{Julius Kunze}{ucl}
\icmlauthor{Louis Kirsch}{ucl,idsia}
\icmlauthor{Hippolyt Ritter}{ucl}
\icmlauthor{David Barber}{ucl,ati}
\end{icmlauthorlist}

\icmlaffiliation{ucl}{University College London}
\icmlaffiliation{ati}{Alan Turing Institute}
\icmlaffiliation{idsia}{IDSIA, The Swiss AI Lab (USI \& SUPSI)}

\icmlcorrespondingauthor{Julius Kunze}{juliuskunze@gmail.com}

\icmlkeywords{Machine Learning, ICML}

\vskip 0.3in
]}

\date{\today}

\usepackage{xcolor}

\newcommand\makeabstract{\begin{abstract} Variational inference with a factorized Gaussian posterior estimate is a widely used approach for learning parameters and hidden variables. Empirically, a regularizing effect can be observed that is poorly understood. In this work, we show how mean field inference improves generalization by limiting mutual information between learned parameters and the data through noise. We quantify a maximum capacity when the posterior variance is either fixed or learned and connect it to generalization error, even when the KL-divergence in the objective is rescaled. Our experiments demonstrate that bounding information between parameters and data effectively regularizes neural networks on both supervised and unsupervised tasks. \end{abstract}}

\usepackage{amsmath,amsfonts,bm}

\def\eqref#1{equation~\ref{#1}}

\def\1{\bm{1}}

\DeclareMathAlphabet{\mathsfit}{\encodingdefault}{\sfdefault}{m}{sl}
\SetMathAlphabet{\mathsfit}{bold}{\encodingdefault}{\sfdefault}{bx}{n}

\newcommand{\KL}{D_{\mathrm{KL}}}

\newcommand \p[1] {\left(#1\right)}

\newcommand \expect {\mathbb{E}}
\newcommand \gauss[1] {\operatorname{\mathcal{N}}\p{#1}}

\newcommand{\D}[2]{\KL\p{{#1}||{#2}}}

\newcommand{\const}{\text{const.}}
\newcommand{\bits}{\text{ bits}}

\newcommand{\corrupt}[1]{\tilde{#1}}

\newcommand{\modelparam}{\theta}
\newcommand{\noisymodelparam}{\corrupt{\modelparam}}
\newcommand{\meanparam}{\mu}
\newcommand{\sdparam}{\sigma}
\newcommand{\varparam}{\sdparam^2}
\newcommand{\params}{\meanparam, \varparam}
\newcommand{\modelparams}{\modelparam, \varparam}

\newcommand{\modelparami}{\modelparam_i}
\newcommand{\noisymodelparami}{\noisymodelparam_i}
\newcommand{\meanparami}{\meanparam_i}
\newcommand{\varparami}{\varparam_i}

\newcommand{\modelparamsi}{\modelparami, \varparami}

\newcommand{\sdprior}{\sigma_p}
\newcommand{\varprior}{\sdprior^2}

\newcommand \orig[1] {#1}
\newcommand \adapt[1] {#1'}

\newcommand*\diff{\mathop{}\!\mathrm{d}}

\newcommand\tikzscale[1]{\scalebox{.85}{#1}}

\usepackage[accepted]{icml2019}

\begin{document}
    \icmlheader
    
    \printAffiliationsAndNotice{}
    \makeabstract
    
    \section{Introduction}

Bayesian machine learning is a popular framework for dealing with uncertainty in a principled way by integrating over model parameters rather than finding point estimates \citep{bishop,barber,zoubin}.
Unfortunately, exact inference is usually not feasible due to the intractable normalization constant of the posterior.
A popular alternative is variational inference \citep{wainwright2008graphical}, where a tractable approximate distribution is optimized to resemble the true posterior as closely as possible.
Due to its amenability to stochastic gradient descent \citep{hoffman2013stochastic,kingma2013auto,titsias2014doubly,rezende2015variational}, variational inference is scalable to large models and datasets.

The most common choice for the variational posterior is a factorized Gaussian.
Outside of Bayesian inference, parameter noise has been found to be an effective regularizer \citep{graves2013speech,plappert2018parameter,fortunato2018noisy}, e.g. for training neural networks.
In combination with $L2$-regularization, additive Gaussian parameter noise corresponds to variational inference with a Gaussian approximate posterior with fixed variance.
Interestingly, it has been observed  that flexible posteriors can perform worse than simple ones \citep{turner2011two,trippe2018overpruning,braithwaite2018bounded,shu2018amortized}.

Variational inference follows the Minimum Description Length (MDL) principle \citep{rissanen1978modeling,rissanen1983universal,hinton93keeping}, a formalization of Occam's Razor.
Loosely speaking, it states that of two models describing the data equally well, the `simpler' one should be preferred.
However, MDL is only an objective for compressing the training data and the model, and makes no formal statement about generalization to unseen data.
Yet, generalization to new data is a key property of a machine learning algorithm.

Recent work \citep{xu2017information,bu2019tightening,bassily2018learners,russo2015much} has proposed upper bounds on the generalization error as a function of the mutual information between model parameters and training data. It states that the gap between train and test error can be reduced by limiting the mutual information.
However, to the best of our knowledge these bounds and specific inference methods have so far not been linked.

In this work, we show that Gaussian mean field inference in models with Gaussian priors can be reinterpreted as point estimation in corresponding noisy models.
This leads to an upper bound on the mutual information between model parameters and data through the data processing inequality.
Our result holds for both supervised and unsupervised models.
We discuss the connection to generalization bounds from \citet{xu2017information} and \citet{bu2019tightening}, suggesting that Gaussian mean field aids generalization.
In our experiments, we show that limiting model capacity via mutual information is an effective measure of regularization, further supporting our theoretical framework.
    \section{Regularization through Mean Field}
\label{sec:theory}

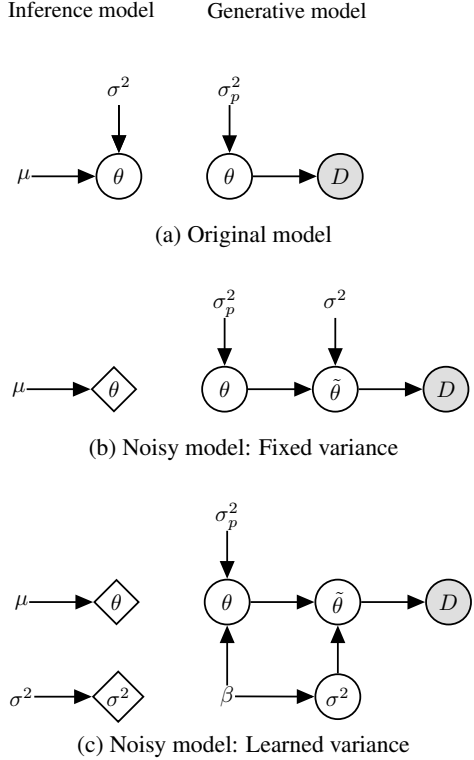
\begin{figure}[ht]
    \centering
    \subfloat[Original model]{
        \centering
        \tikzscale{\begin{tikzpicture}[thick]
    \node[const] (mu) {$\meanparam$};
    \node[latent, right=of mu] (thetainf) {$\modelparam$};
    \node[latent, right=of thetainf] (thetamodel) {$\modelparam$};
    \node[obs, right=of thetamodel] (D) {$D$};
    \node[const, above=.75cm of thetamodel] (sigmap) {$\varprior$};
    \node[const, above=.75cm of thetainf] (sigma) {$\varparam_{\phantom{p}}$};
    
    \node[latent, right=of D, draw=none] (phantom) {};
    
    \node[const, below=.4cm of mu] (placeholderbottom) {};
    
    \node[const, above=2.35cm of mu, xshift=.9cm] (inf) {Inference model};
    \node[const, above=2.1cm of thetamodel, xshift=.9cm] (gen) {Generative model};
    
    \edge{sigma}{thetainf}
    \edge{sigmap}{thetamodel}
    \edge{mu,sigma}{thetainf}
    \edge{thetamodel}{D}
\end{tikzpicture}}
        \label{fig:mean-field-original}
    }\\[0.5cm]
    \subfloat[Noisy model: Fixed variance]{
        \centering
        \tikzscale{\begin{tikzpicture}[thick]
    \node[const] (mus) {$\meanparam$};
    \node[det, right=of mus] (thetainf) {$\modelparam$};
    \node[latent, right=of thetainf] (thetamodel) {$\modelparam$};
    \node[latent, right=of thetamodel] (thetanoisy) {$\noisymodelparam$};
    \node[obs, right=of thetanoisy] (D) {$D$};
    \node[const, above=.75cm of thetamodel] (sigmap) {$\varprior$};
    \node[const, above=.75cm of thetanoisy] (var) {$\varparam_{\phantom{p}}$};

    \node[const, below=.4cm of mu] (placeholderbottom) {};
    
    \edge{mus}{thetainf}
    \edge{sigmap}{thetamodel}
    \edge{thetamodel,var}{thetanoisy}
    \edge{thetanoisy}{D}

\end{tikzpicture}}
        \label{fig:fixed-variance-NIB}
    }\\[0.5cm]
    \subfloat[Noisy model: Learned variance]{
        \centering
        \tikzscale{\begin{tikzpicture}[thick]
    \node[const] (mus) {$\meanparam$};
    \node[det, right=of mus] (thetainf) {$\modelparam$};
    \node[latent, right=of thetainf] (thetamodel) {$\modelparam$};
    \node[latent, right=of thetamodel] (thetanoisy) {$\noisymodelparam$};
    \node[obs, right=of thetanoisy] (D) {$D$};
    \node[const, above=.75cm of thetamodel] (sigmap) {$\varprior$};

    \node[const, below=.4cm of mu] (placeholderbottom) {};
    
    \node[const, below=0.94cm of thetamodel] (beta) {$\beta$};

    \node[const, below=1.2cm of mus] (sigmas) {$\varparam$};
    \node[det, right=.9cm of sigmas] (sigmainf) {$\varparam$};
    \node[latent, below=0.75cm of thetanoisy] (var) {$\varparam$};
    
    \edge{beta}{thetamodel, var}
    \edge{mus}{thetainf}
    \edge{sigmas}{sigmainf}
    \edge{sigmap}{thetamodel}
    \edge{thetamodel,var}{thetanoisy}
    \edge{thetanoisy}{D}

\end{tikzpicture}}
        \label{fig:learned-variance-NIB}
    }
    \caption{Gaussian mean field inference on model parameters $\modelparam$ with a Gaussian prior (a) can be reinterpreted as optimizing a point estimate on a model with injected noise both when variance is fixed (b) and learned (c). For the latter case, we show this for the more general case where the complexity term in the objective is scaled by a constant $\beta>0$, with $\beta=1$ recovering variational inference.}
    \label{fig:mean-field}
\end{figure}

In our derivation, we denote a generic model as $p(\modelparam, D) = p(\modelparam) p(D \mid \modelparam)$  with unobserved variables $\theta$ and data $D$. We refer to $\theta$ as the model parameters, however in latent variable models $\theta$ can also include the per-data point latents. The model consists of a prior $p(\modelparam)$ and a likelihood $p(D \mid \modelparam)$. Ideally, one would like to find the posterior $p(\modelparam \mid D) = p(D \mid \modelparam) p(\modelparam) / Z$, where $Z = \int p(D \mid \modelparam) p(\modelparam) d\modelparam$ is the normalizer. However, calculating $Z$ is typically intractable. Variational inference finds an approximation by maximizing the evidence lower bound (ELBO)
\begin{align}
\begin{split}
\log p(D) &\geq \log p(D) - \D{q(\modelparam)}{p(\modelparam \mid D)}\\
     &= \expect_{\orig{q}(\theta)} \log \orig{p}(D \mid  \theta) - \D{\orig{q}(\theta)}{\orig{p}(\theta)}
\end{split}
\label{eq:free-energy}
\end{align}
w.r.t. the approximate posterior $q(\modelparam)$.
Our focus in this work lies on Gaussian mean field inference, so $q$ is a fully factorized normal distribution with a learnable mean $\meanparam$ and variance $\varparam$.
The prior is also chosen to be component-wise independent $p(\modelparam)=\gauss{0, \varprior I}$. The generative and inference models for this setting are shown in \autoref{fig:mean-field-original}.

\subsection{Fixed-Variance Gaussian Mean Field Inference}
\label{sec:fixed-variance}

When the variance $\varparam$ of the approximate posterior is fixed to some constant, the ELBO can be written as
\begin{align}
\label{eq:fixed-original}
\expect_{\modelparam \sim \gauss{\meanparam, \varparam I}} \log \orig{p}(D \mid \modelparam) - \frac{1}{2\varprior}\sum_i \meanparami^2 + \const
\end{align}
which is optimized with respect to $\meanparam$. We use $i \in \{1, \ldots, K\}$ to denote the parameter index.

To show how Gaussian mean field implicitly limits learned information, we extend the model with a noisy version of the parameters $\noisymodelparam \sim p(\noisymodelparam  \mid  \modelparam)$ and let the likelihood depend on those noisy parameters.
We choose the noise distribution to be the same as the inference distribution for the original model and find a lower bound on the log-joint of the noisy model.
This leads to the same objective as mean-field variational inference in the original model.

Specifically, we define the noisy model $\adapt{p}(\modelparam, \noisymodelparam, D)=\adapt{p}(\modelparam)\adapt{p}(\noisymodelparam \mid \modelparam)\adapt{p}(D \mid \noisymodelparam)$ as visualized in \autoref{fig:fixed-variance-NIB}.
We use $\adapt{p}$ to emphasize the distinction between distributions of the modified noisy model and the original one.
As in the original model, $\modelparam \sim \gauss{0, \sigma_p^2 I}$ represents the parameters (with same prior), i.e. $p(\modelparam) = \adapt{p}(\modelparam)$.
We denote the noisy parameters as $\noisymodelparam \sim \gauss{\modelparam, \varparam}$.
The likelihood remains unchanged, i.e. $\adapt{p}(D \mid \noisymodelparam)=\orig{p}(D \mid \theta)$, except that it now depends on the noisy parameters instead of the `clean' ones.

We now show that maximizing a lower bound on the log joint probability of the noisy model results in an identical objective as for variational inference in the clean model
\begin{align}
    &\log \adapt{p}(D, \modelparam) \\
    =& \log \int \adapt{p}(D  \mid  \noisymodelparam) \adapt{p}(\noisymodelparam  \mid  \modelparam) \diff\noisymodelparam  + \log \adapt{p}(\modelparam) \\
    \geq& \expect_{\noisymodelparam \sim \gauss{\theta, \varparam I}} \log \adapt{p}(D \mid \noisymodelparam) - \frac{1}{2\varprior}\sum_i \modelparami^2 + \const \label{eq:noisy-lower-bound}\\
    =& \expect_{\noisymodelparam \sim \gauss{\meanparam, \varparam I}} \log \adapt{p}(D \mid \noisymodelparam) - \frac{1}{2\varprior}\sum_i \meanparami^2 + \const \label{eq:fixed-variance-noisy-objective}
\end{align}
where \autoref{eq:noisy-lower-bound} follows from Jensen's inequality as in \autoref{eq:free-energy}.
In the final equation we have replaced $\modelparam$ with $\meanparam$ (which is simply a change of names since we are maximizing the objective over this free variable) to emphasize that the objective functions are identical.

Since $D$ is independent of $\modelparam$ given $\noisymodelparam$, the joint $p(\modelparam,\noisymodelparam,D)$ forms a Markov chain and the data processing inequality \citep{cover2012elements} limits the mutual information $I(D, \theta)$ between learned parameters and data through
\begin{align}
\label{eq:NIB-capacity}
I(D, \theta) \leq I(\corrupt{\theta}, \theta)
\end{align}
The upper bound is given by
\begin{align}
\label{eq:fixed-capacity}
I(\noisymodelparam, \modelparam) &= H(\noisymodelparam) - H(\noisymodelparam \mid \modelparam) = \frac{K}{2} \log \p{1 + \frac{\varprior}{\varparam}}
\end{align}
where $K$ denotes the number of parameters. Here, we exploit that $\modelparam$ and $\noisymodelparam \mid \modelparam$ are Gaussian with $H(\noisymodelparam)=\frac{K}{2} \log 2\pi e \p{\varparam + \varprior}$ and  $H(\noisymodelparam \mid \modelparam)=\frac{K}{2} \log 2\pi e \sigma^2$. This quantity is known as the capacity of channels with Gaussian noise in signal processing \citep{cover2012elements}. Intuitively, a high prior variance $\varprior$ corresponds to a large capacity, while a high noise variance $\varparam$ reduces it. Any desired capacity can be achieved by simply adjusting the signal-to-noise ratio $\varprior / \varparam$.

\subsection{Generalization Error vs. Limited Information}
\label{sec:generalization-error}

Intuitively, we characterize overfitting as learning too much information about the training data, suggesting that limiting the amount of information extracted from the training data into the hypothesis should improve generalization.
This idea has recently been formalized by \citet{xu2017information,bu2019tightening,bassily2018learners,russo2015much} showing that limiting mutual information between data and learned parameters bounds the expected generalization error under certain assumptions. 

Specifically, their work characterizes the following process: Assume that our training dataset is sampled from a true distribution $p_t(D)$. Based on this training set, a learning algorithm subsequently returns a distribution over hypotheses given by $p_t(\modelparam \mid D)$. The process defines a mutual information $I_t(D, \modelparam)$ on the joint distribution $p_t(D, \modelparam) = p_t(D) p_t(\modelparam \mid D)$. Under certain assumptions on the loss function, \citet{xu2017information} derive a bound on the generalization error of the learning algorithm in expectation over this sampling process. \citet{bu2019tightening} relax the condition on the loss and prove applicability to a simple estimation algorithm involving $L2$-loss.

Exact Bayesian inference returns the true posterior $p(\theta  \mid  D)$ on a model $p(\theta, D)$. The theorem then states that a bound on $I(D, \theta)$ limits the expected generalization error as described in \citet{bu2019tightening} if the model captures the nature of the generating process in the marginal $p(D) = \int \diff \theta p(\theta) p(D \mid \theta)$.
This is a common assumption necessary to justify any (variational) Bayesian approach.

Exact inference is intractable on deep models, and instead, one typically learns variational or point estimates for the posterior.
That is also true for the objective on the noisy model above, where we only used a point estimate as given by \autoref{eq:fixed-variance-noisy-objective}.
Therefore, the assumption of exact inference is not met.
Yet, we believe that those bounds motivate the expectation that variational inference aids generalization by limiting the learned information.
If we performed exact inference on the noisy model in the last section, the given mutual information would imply a bound on generalization error as implied by \citet{xu2017information} and \citet{bu2019tightening}.
Therefore, we are optimistic that the gap between variational inference and those generalization bounds can be closed either by performing more accurate inference in the noisy model or by taking the dynamics of the training algorithm into account when bounding mutual information (see \autoref{sec:learning-dynamics} for further discussion).


\subsection{Learned-Variance Gaussian Mean Field Inference}
\label{sec:learned-variance}

The variance in Gaussian mean field inference is typically learned for each parameter \citep{kingma2015variational,rezende2015variational,blundell2015weight}.
Similar to when the variance in the approximate posterior is fixed, one can obtain a capacity constraint.
This is the case even for a generalization of the objective from \autoref{eq:free-energy} where the KL-term is scaled by some factor $\beta>0$.\footnote{\citet{higgins2016beta} propose using $\beta > 1$ to learn `disentangled' representations in variational autoencoders. Further, $\beta$ is commonly annealed from $0$ to $1$ for expressive models (e.g. \citet{bowman2015generating,blundell2015weight,sonderby2016ladder}).}
In the following, we quantify a general capacity depending on $\beta$, where $\beta=1$ recovers the standard variational objective.
For notational simplicity, we here assume a prior variance of $\sigma_p^2=1$. It is straight-forward to adapt the derivation to the general case.

In this case, the objective can be written as
\begin{align}
\label{eq:beta-original}
\begin{split}
&\expect_{\modelparam \sim \gauss{\params}} \log \orig{p}(D \mid \modelparam) \\
&+ \frac{\beta}{2} \sum_i \p{\log \varparami - \varparami - \meanparami^2 - 1}
\end{split}
\end{align}
where now both $\meanparam$ and $\varparam$ represent learned vectors, and $\gauss{\params}$ denotes a variable composed of pairwise independent Gaussian components with means and variances given by the elements of $\meanparam$ and $\varparam$.

Similar to the previous section, we show a lower bound on the log-joint of a new noisy model to be identical to \autoref{eq:beta-original}.
Specifically, we define the noisy model $\adapt{p}(\modelparam, \varparam, \noisymodelparam, D)=\adapt{p}(\modelparam)\adapt{p}(\varparam)\adapt{p}(\noisymodelparam \mid \modelparam, \varparam)\adapt{p}(D \mid \noisymodelparam)$ (\autoref{fig:learned-variance-NIB}), with independent priors
$\modelparami \sim \gauss{0, \frac{1}{\beta}}$ and
$\varparami \sim \Gamma \p{\frac{\beta}{2}+1, \frac{\beta}{2}}$ where $\Gamma(\cdot, \cdot)$ denotes the Gamma distribution.
As previously done \autoref{sec:fixed-variance}, we define the noise-injected parameters as $\noisymodelparam \sim \gauss{\modelparam, \varparam}$ and likelihood as $\adapt{p}(D \mid \noisymodelparam)=\orig{p}(D \mid \theta)$.

The priors are chosen so that with Jensen's inequality, we find a lower bound on the log-joint probability of this model that recovers the objective from \autoref{eq:beta-original}
\begin{align}
\begin{split}
    &\log \adapt{p}(D, \modelparam, \varparam) \\
    =& \log \int \adapt{p}(D  \mid  \noisymodelparam) \adapt{p}(\noisymodelparam  \mid  \modelparam, \varparam) \diff\noisymodelparam  + \log \adapt{p}(\modelparam) + \log \adapt{p}(\varparam) \\
    \geq& \expect_{\noisymodelparam \sim \gauss{\modelparams}} \log \adapt{p}(D \mid \noisymodelparam) + \sum_i \p{\log \adapt{p}(\modelparami) + \log \adapt{p}(\varparami)}\\
    =& \expect_{\noisymodelparam \sim \gauss{\params}} \log \adapt{p}(D \mid \noisymodelparam) \\
    &+ \frac{\beta}{2} \sum_i \p{\log \varparami - \varparami - \meanparami^2} + \const
\end{split}
\end{align}

\begin{figure}
    \centering
    \includegraphics[width=\linewidth]{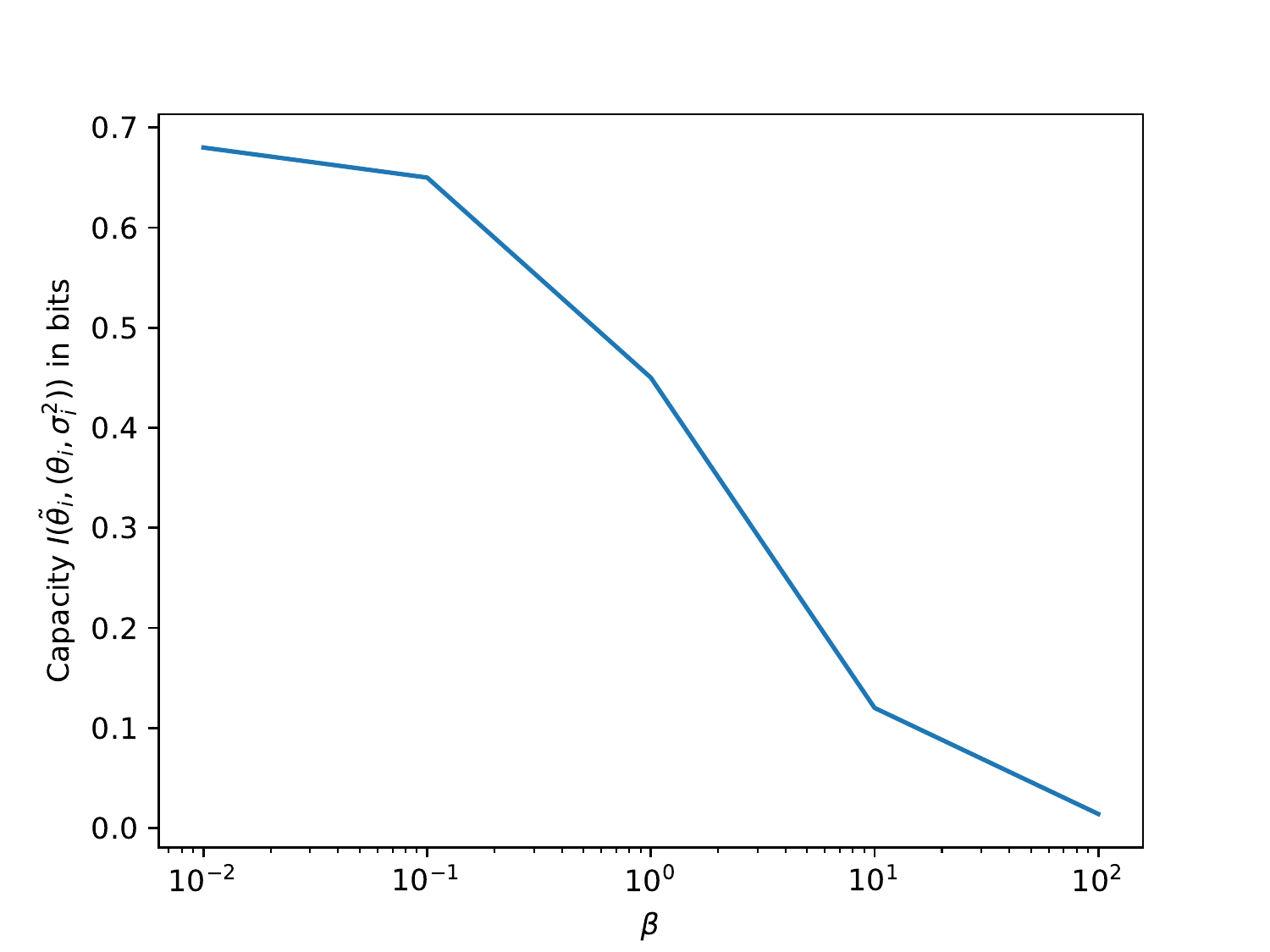}
    \caption{Relationship between $\beta$ and capacity $I(\noisymodelparami, (\modelparamsi))$ per parameter dimension in Gaussian mean field inference with learned variance and complexity term scaled by $\beta>0$.
    }
    \label{fig:beta-vae-capacity}
\end{figure}

In the noisy model, the data processing inequality and the independence of dimensions implies a bound
\begin{align}
    I(D, (\modelparams)) \leq I(\noisymodelparam, (\modelparams)) = \sum_i I(\noisymodelparami, (\modelparamsi)) 
\end{align}
where the capacity $I(\noisymodelparami, (\modelparamsi))$ per dimension is derived in \autoref{sec:learned-variance-mi-derivation}.
\autoref{fig:beta-vae-capacity} shows numerical results for various values of $\beta$.
Standard variational inference ($\beta=1$) results in a capacity of $0.45 \bits$ per dimension.
We observe that higher $\beta$ corresponds to smaller capacity, which is given by the mutual information $I(\noisymodelparami, (\modelparamsi))$ between our new latent $(\modelparamsi)$ and $\noisymodelparami$.
This formalizes the intuition that a higher weight of the complexity term in our objective increases regularization by decreasing a limit on the capacity.

\subsection{Supervised and Unsupervised Learning}\label{sec:supervised_unsupervised}

The above derivations apply to any learning algorithm that is purely trained with Gaussian mean-field inference. This covers supervised and unsupervised tasks. 

In supervised learning, the training data typically consists of pairs of inputs and labels, and a loss is assigned to each pair that depends on the trained model, e.g. neural network parameters. When all parameters are learned with one of the discussed mean-field methods, the given bounds apply.

The derivation also comprises unsupervised methods with per-data latents and even amortized inference such as VAEs \citep{kingma2015variational,rezende2015variational}, again as long as all learned variables are learned via Gaussian mean field inference.
While this might be helpful to find generalizing representations, the focus of the experiments is on validating the generalizing behavior of mean field variational Bayes on neural network parameters for  overfitting regimes, namely small dataset and complex models.

\subsection{Flexible Variational Distributions}
\label{sec:flexible-variational}

The objective function for variational inference is maximized when the approximate posterior is equal to the true one.
This motivates the development of flexible families of posterior distributions \citep{rezende2015variational,kingma2016improved,salimans2015markov,ranganath2016operator,huszar2017variational,chen2018neural,vertes2018flexible,burda2015importance,cremer2017reinterpreting}.
In the case of exact inference, a bound on generalization as discussed in \autoref{sec:generalization-error} only applies if the model itself has finite mutual information between data and parameters. However, estimating mutual information is generally a hard problem, particularly in high-dimensional, non-linear models. This makes it hard to state a generic bound, which is why we focus on the case of Gaussian mean field inference.
    \section{Related Work}
\label{sec:related-work}

\paragraph{Regularization in Neural Networks} Gaussian mean field is intimately related with other popular regularization approaches in deep learning: 
As apparent from \autoref{eq:fixed-variance-noisy-objective}, fixed-variance Gaussian mean field applied to training neural network weights is equivalent to \textit{L}2-regularization (weight decay) combined with Gaussian parameter noise \citep{graves2013speech,plappert2018parameter,fortunato2018noisy} on all network weights.
\citet{molchanov2017variational} shows that additive parameter noise results in multiplicative noise on the unit activations.
The resulting dependencies between noise components on the layer output can be ignored without significantly changing empirical results \cite{Wang2013fast}. This is in turn equivalent to scaled Gaussian dropout \citep{kingma2015variational}.

\paragraph{Information Bottlenecks} The Information Bottleneck principle by \citet{tishby2000information,shamir2010learning} aims to find a representation $Z$ of some input $X$ that is most useful to predict an output $Y$.
For this purpose, the objective is to maximize the amount of information $I(Y, Z)$ the representation contains about the output under a bounded amount of information $I(X, Z)$ about the input
\begin{align}\label{eq:IB-constrained}
    \max_{I(X, Z) < C} I(Y, Z)
\end{align}
They describe a training procedure using the softly constrained objective
\begin{align}\label{eq:IB}
    \min \mathcal{L}_{IB} = \min I(X, Z) - \beta I(Y, Z)
\end{align}
where $\beta>0$ controls the trade-off.

\citet{alemi2016deep} suggest a variational approximation for this objective.
For the task of reconstruction, where labels $Y$ are identical to inputs $X$, this results exactly in the $\beta$-VAE objective \citep{achille2017emergence, alemi2018fixing}.
This is in accordance with our result from \autoref{sec:learned-variance} that there is a maximum capacity per latent dimension that decreases for higher $\beta$.
Setting $\beta>1$, as suggested by \citet{higgins2016beta} for obtaining disentangled representations, corresponds to lower capacity per latent component than achieved by standard variational inference.

Both \citet{tishby2000information} and \citet{higgins2016beta} introduce $\beta$ as a trade-off parameter without a quantitative interpretation. 
With our information-theoretic perspective, we quantify the implied capacity and provide a link to the generalization error.
Further, both methods are concerned with the information in the latent representation. 
They do not limit the mutual information with the model parameters, leaving them vulnerable to model overfitting under our theoretical assumptions.
We experimentally validate this vulnerability and explore the effect of filling this gap by applying Gaussian mean field inference to the model parameters.

\paragraph{Information Estimation with Neural Networks} Multiple recent techniques \citep{belghazi2018mine,oord2018representation,hjelm2018learning} propose the use of neural networks for obtaining a \emph{lower} bound on the mutual information.
This is useful in settings when we want to \emph{maximize} mutual information, e.g. between the data and a lower-dimensional representation. In contrast, we show that Gaussian variational inference on variables with a Gaussian prior implicitly places an \emph{upper} bound on the mutual information between these variables and the data, and explore its regularizing effect.
    \section{Experiments}\label{sec:experiments}

In this section, we analyze the implications of applying Gaussian mean field inference of fixed scale to the model parameters in the supervised and unsupervised context.
Our theoretical results suggest that varying the capacity will affect the generalization capability and we show this effect on small data regimes and how it changes with the training set size.
Furthermore, we investigate whether capacity is the only predictor for generalization or whether varying priors and architectures also have an effect.
Finally, we demonstrate qualitatively how the capacity bounds are reflected in fashion MNIST reconstruction.

\subsection{Supervised Learning}\label{sec:gen_supervised}

\begin{figure}
    \centering
    \includegraphics[width=\linewidth]{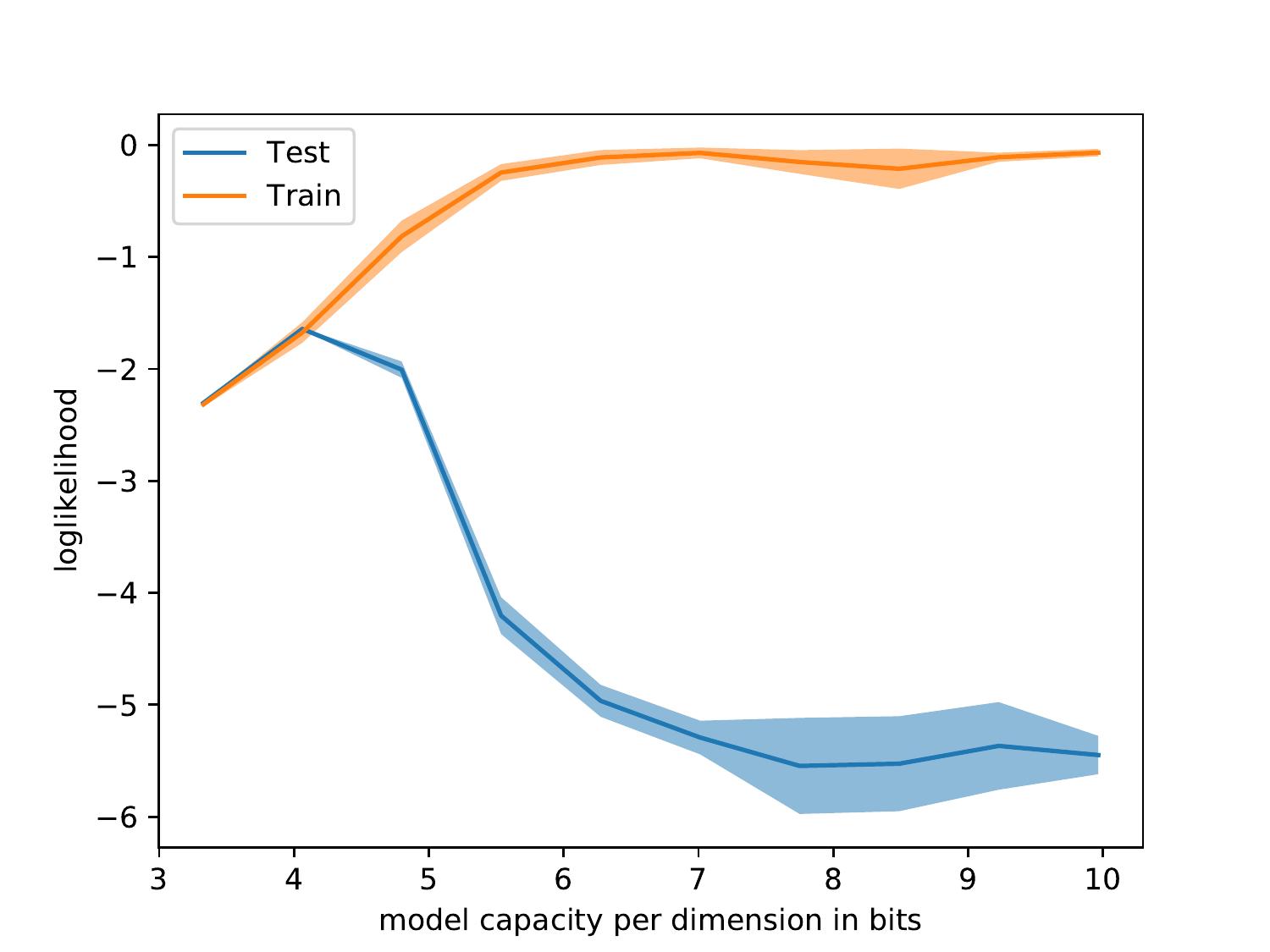}
    \caption{Classifying CIFAR10 with varying model capacities.
    Large capacities lead to overfitting while small capacities drown the signal in noise.
    Each configuration has been evaluated 5 times; mean and standard deviation are displayed.}
    \label{fig:supervised_generalization}
\end{figure}

We begin with a supervised classification task on the CIFAR10 dataset, training only on a subset of the first 5000 samples. 
We use 6 3x3 convolutional layers with 128 channels each followed by a ReLU activation function, every second of which implements striding 2 to reduce the input dimensionality.
Finally, the last layer is a linear projection which parameterizes a categorical distribution.
The capacity of each parameter in this network is set to specific values given by \autoref{eq:fixed-capacity}.

\autoref{fig:supervised_generalization} shows that decreasing the model capacity per dimension (by increasing the noise) reduces the training log-likelihood and increases the test loglikelihood until both of them meet at an optimal capacity.
One can observe that very small capacities lead to a signal that is too noisy and good predictions are no longer possible.
In short, regimes of underfitting and overfitting are generated depending on the capacity.

\subsection{Unsupervised Learning}\label{sec:gen_unsupervised}

We now evaluate the regularizing effect of fixed-scale Gaussian mean field inference in an unsupervised setting for MNIST image reconstruction.
Therefore, we use a VAE~\citep{kingma2013auto} with 2 latent dimensions and a 3-layer neural network parameterizing the conditional factorized Gaussian distribution.
As usual, it is trained using the free energy objective, but different from the original work, we also use Gaussian mean field inference for the model parameters.
Again, we use a small training set of 200 examples for the following experiments if not denoted otherwise.

\begin{figure*}[ht]
    \centering
    \subfloat[The test ELBO is not invariant when varying the prior on the model parameters. Neverthless, the first increasing and then decreasing trend when changing the capacity remains.]{
        \includegraphics[width=0.46\textwidth]{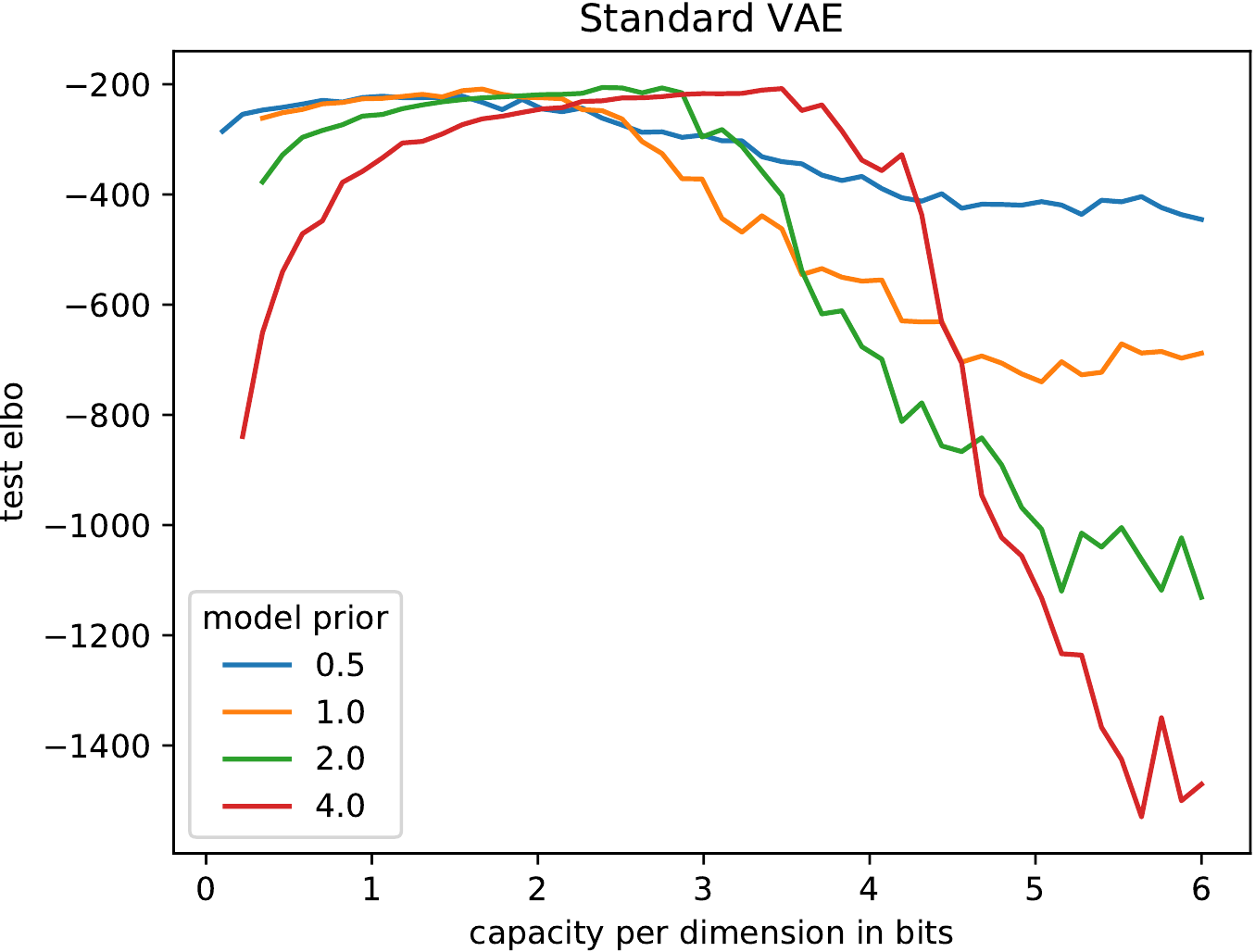}
        \label{fig:unsupservised_gen}
    }
    \quad
    \subfloat[Using an improper prior, similar to just using Gaussian dropout on the weights, leads to an accelerated decrease of generalization for smaller noise scales.]{
         \includegraphics[width=0.46\textwidth]{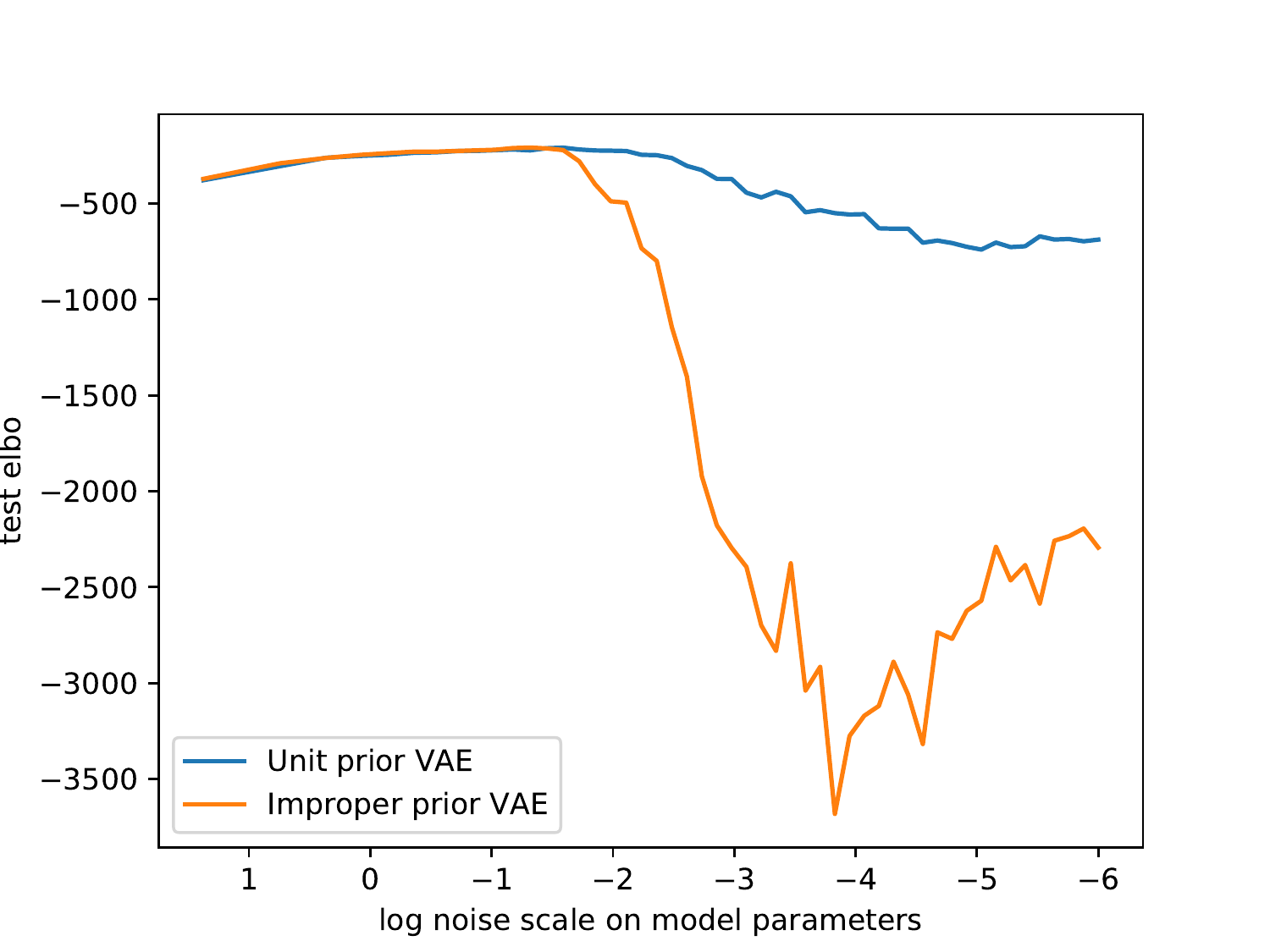}
        \label{fig:dropping_prior}
    }
    \caption{MNIST test reconstruction with a VAE training on 200 samples for various priors and capacities.}
\end{figure*}

\paragraph{Varying Model Capacity and Priors} In our first experiment, we analyze generalization by inspecting the test ELBO when varying the model capacity which can be seen in \autoref{fig:unsupservised_gen}.
Similar to the supervised case, we can observe that there is a certain model capacity range that explains the data very well while less or more capacity results in noise drowning and overfitting respectively.
In the same figure, we also investigated whether the information-theoretic model capacity can predict generalization independently of the specific prior distribution.
Since we merely state an upper bound on mutual information in \autoref{sec:fixed-variance}, the prior may have an effect in practice which cannot be explained only by the capacity.
\autoref{fig:unsupservised_gen} shows that indeed, while the general behavior remains the same for different model priors, the generalization error is not entirely independent.
Furthermore, the observation that all curves descend with larger capacities, for all priors, suggests that weight decay \citep{krogh1992simple} of fixed scale without parameters noise is not sufficient to regularize arbitrarily large networks.
In \autoref{fig:dropping_prior} we investigated the extreme case of dropping the prior entirely and switching to maximum-likelihood learning instead by using an improper uniform prior.
This approach recovers Gaussian dropout~\cite{srivastava2014dropout,kingma2015variational}.
Dropping the prior sets the bottleneck capacity to infinity and should lead to worse generalization.
Comparing the test ELBO of this Gaussian dropout variant to the original Gaussian mean field inference in \autoref{fig:dropping_prior} confirms this result for larger capacities.
For larger noise scales, generalization is still working well, a result that is not explained in our information-theoretic framework, but plausible due to the deployed limited architecture.

\begin{figure*}[ht]
    \centering
    \subfloat[Varying the number of samples.
    Depending on the size of the dataset higher capacities of the model are required to fit all the datapoints.]{
        \includegraphics[width=0.46\textwidth]{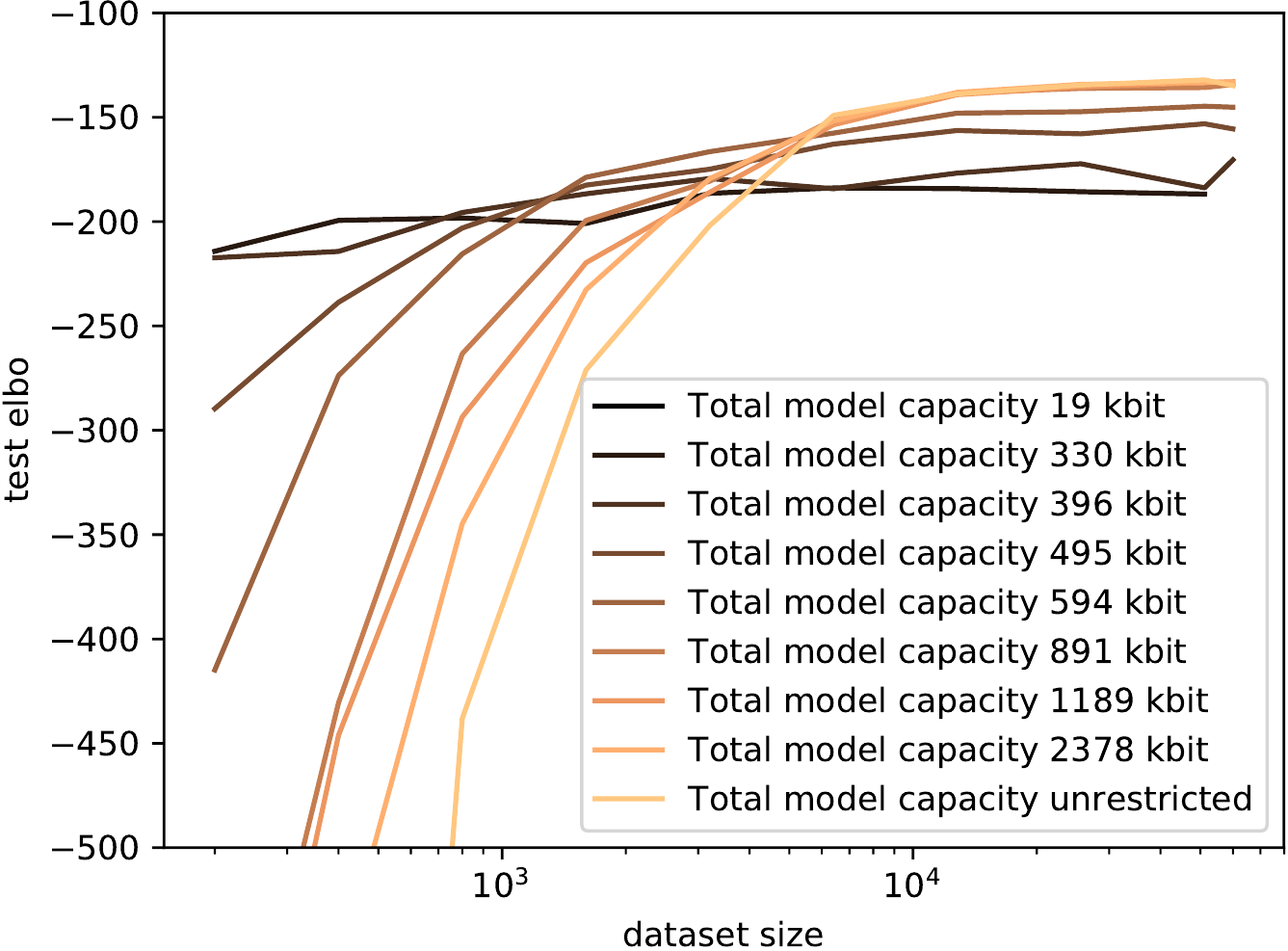}
        \label{fig:vary_dataset_size}
    }
    \quad
    \subfloat[Varying architecture.
        Overfitting is not getting worse for more layers if capacity is low enough.
        More layers do overfit only for higher capacities.]{
        \includegraphics[width=0.46\textwidth]{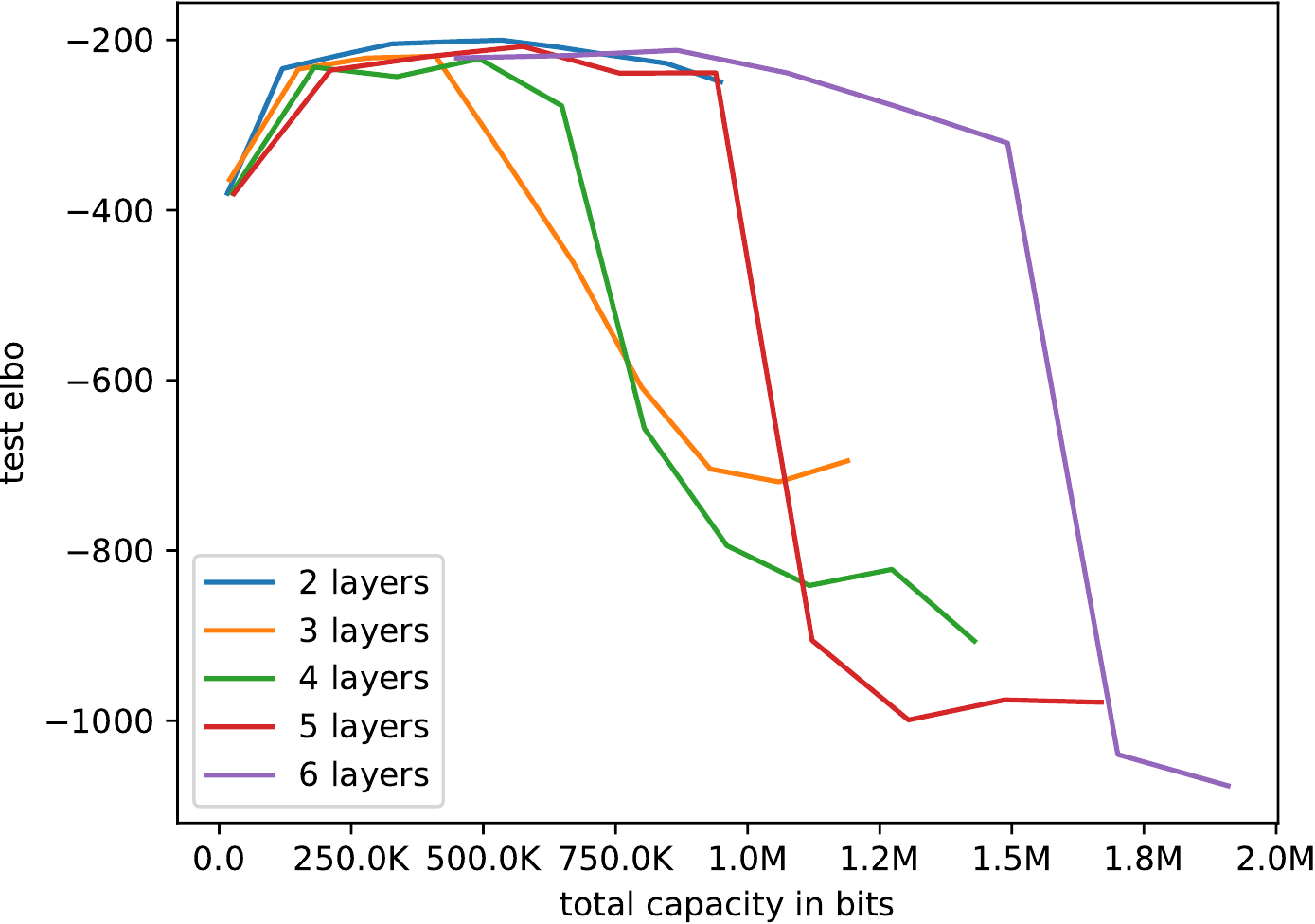}
        \label{fig:vary_depth}
    }
    \caption{MNIST test reconstruction with a VAE; training on varying dataset sizes, architectures, and model capacities.}\label{fig:}
\end{figure*}

\paragraph{Varying Training Set Size} \autoref{fig:vary_dataset_size} shows how limiting the capacity affects the test ELBO for varying amounts of training data.
Models with very small capacity extract less information from the data into the model, thus yielding a good test ELBO somewhat independent of the dataset size.
This is visible as a graph that ascends very little with more training data (e.g. total model capacity 330 kbits).
Note that we here report the capacity of the entire model, which is the sum of the capacities for each parameter.
In order to improve the test ELBO, more information from the data has to be extracted into the model.
But clearly, this leads to non-generalizing information being extracted when the dataset is small, leading to overfitting.
Only for larger datasets the extracted information generalizes.
This is visible as a strongly ascending test ELBO with larger dataset sizes and bad generalization for small datasets.
We can therefore conclude that the information bottleneck needs to be chosen based on the amount of data that is available.
Intuitively, when more information is available, more information should be extracted into the model.

\paragraph{Varying Model Size} Furthermore, we inspected how the size of the model (here in terms of number of layers) affects generalization in \autoref{fig:vary_depth}.
Similar to varying the prior distribution, we are interested in how well the total capacity predicts generalization and the role the architecture plays.
It can be observed that larger networks are more resilient to larger total capacities before they start overfitting.
This indicates that the total capacity is less important than the individual capacity (i.e. noise) per parameter.
Nevertheless, larger networks are more prone to overfitting for very large model capacities.
This makes sense as their functional form is less constrained, an aspect that is not captured by our theory.

\begin{figure*}
    \centering
    \includegraphics[page=1,width=\textwidth]{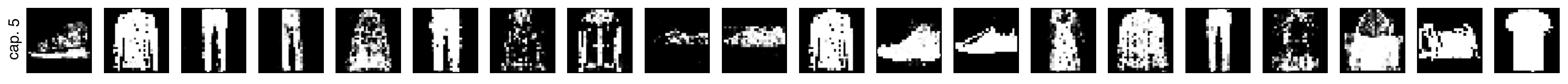}
    \includegraphics[page=2,width=\textwidth]{figures/reconstruction.pdf}
    \includegraphics[page=5,width=\textwidth]{figures/reconstruction.pdf}
    \caption{Test reconstruction means for binarized fashion MNIST trained on 200 samples with per-parameter capacities $5$, $2$ and $1 \bits$ (from top) compared to the true data (bottom).}
    \label{fig:reconstruction-visualizations}
\end{figure*}

\paragraph{Qualitative Reconstruction} Finally, we plot test reconstruction means for the binarized fashion MNIST dataset under the same setup for various capacities in \autoref{fig:reconstruction-visualizations}.
In accordance with previous experiments, we observe that if the capacity is chosen too small, the model is not learning anything useful, while too large capacities result in overconfidence.
This can be observed in most means being close to either 0 or 1.
An intermediate capacity, on the other hand, makes sensible predictions (given that it was trained only on 200 samples) with sensible uncertainty, visible through gray pixels that correspond to high entropy.

    \section{Discussion}

In this section, we discuss how the capacity can be set, as well as the effect of model architecture and learning dynamics.

\subsection{Choosing the Capacity}

We have obtained a new trade-off parameter, the capacity, that has a simple quantitative interpretation: It determines how many bits to maximally extract from the training set.
In contrast, for the $\beta$ parameter introduced in \citet{tishby2000information} and \citet{higgins2016beta} a clear interpretation is not known.
Yet, it may still be hard to set the capacity optimally.
Simple mechanisms such as evaluation on a validation set to determine its value may be used.
We expect that more theoretically rigorous methods could be developed.

Furthermore, in this paper, we have focused on the regularization that Gaussian mean field inference implies on the model parameters.
The same concept is valid for data-dependent latent variables, for instance in VAEs, as discussed in \autoref{sec:supervised_unsupervised}.
In VAEs, Gaussian mean field inference on the latents leads to a restricted latent capacity, but leaves the capacity of the model unbounded.
This leaves VAEs vulnerable to model overfitting, as demonstrated in the experiments, and setting $\beta$ as done in \citet{higgins2016beta} is not sufficient to control complexity.
This motivates the limitation of capacity between the data and both per-datapoint latents and model parameters.
The interaction between the two is an interesting future research direction.

\subsection{Role of Learning Dynamics and Architecture}
\label{sec:learning-dynamics}

As discussed in \autoref{sec:generalization-error}, it is necessary to perform exact inference in the noisy model for the bounds on the generalization error to hold.
However, this assumption is not met.
In practice $p_t(\modelparam \mid D)$ encodes the complete learning algorithm, which in deep learning typically includes parameter initialization and dynamics of the stochastic gradient descent optimization. 

Our experiments confirm the relevance of these other factors:
$L2$-regularization works in practice, even though no noise is added to the parameters. This could be explained by the fact that noise is already implicitly added through stochastic gradient descent \citep{lei2018implicit} or through the output distribution of the network.
Similarly, Gaussian dropout \citep{graves2013speech,plappert2018parameter,fortunato2018noisy} without a prior on the parameters helps generalization. Again, early stopping combined with a finite reach of gradient descent steps effectively shapes a prior of finite variance in the parameter space. 
This could also formalize why the annealing schedule employed by \citet{bowman2015generating,blundell2015weight} and \citet{sonderby2016ladder} is effective.

This observed dependence on other factors suggests that quantifying mutual information $I_t(\modelparam,D)$ of the actual distribution created by the learning dynamics might be a promising approach to explain why neural networks often generalize well on their own.
This idea is in accordance with recent work that links the learning dynamics of small neural networks to generalization behavior \citep{li2018learning}.

On the other hand, the architecture choice also had an influence on generalization, which is expected by our theory since we only formulate a bound on mutual information that is completely agnostic to the actual model choice. Tightening this bound based on the model architecture and output distribution is usually hard, as discussed in \autoref{sec:flexible-variational}, but might be possible.

Another promising direction would be to approximately sample from the exact posterior on network parameters (i.e. as done by \citet{marceau2017natural}), on a capacity-limited architecture, instead of the usual approach of point estimation. In the limit of infinite training time, this would fully realize the discussed bound on the expected generalization error.
    \section{Conclusion}
\label{sec:conclusion}

We have explained the regularizing effects observed in Gaussian mean field approaches from an information-theoretic perspective. The derivation features a capacity that can be naturally interpreted as a limit on the amount of information extracted about the given data by the inferred model. We validated its practicality for both supervised and unsupervised learning.

How this capacity should be set for parameters and latent variables depending on task and data is an interesting direction of research.
We exploited a theoretical link of mutual information and generalization error. While this work is restricted to Gaussian mean field, incorporating the effect of learning dynamics on mutual information in future work might allow understanding why overparameterized neural networks still generalize well to unseen data.
    
    \appendix \section{Capacity in Learned-Variance Gaussian Mean Field Inference}
\label{sec:learned-variance-mi-derivation}

The capacity per dimension for the model discussed in \autoref{sec:learned-variance} is given by
\begin{align}
\begin{split}
\label{eq:beta-vae-capacity}
 & I(\noisymodelparami, (\modelparamsi)) \\
=& H(\noisymodelparami) - H(\noisymodelparami \mid \modelparamsi) \\
=& -\int_{-\infty}^\infty \adapt{p}(\noisymodelparami) \log \adapt{p}(\noisymodelparami) \diff \noisymodelparami \\
&-\int_0^\infty \adapt{p}(\varparami) \frac{1}{2} \log 2 \pi e \varparami \diff \varparami
\end{split}
\end{align}
$\noisymodelparami \mid \modelparamsi \sim \gauss{\modelparamsi}$ with $\modelparami \sim \gauss{0, \frac{1}{\beta}}$ implies $\noisymodelparami \mid \varparami \sim \gauss{0, \varparami + \frac{1}{\beta}}$. Together with $\varparami \sim \Gamma \p{\frac{\beta}{2}+1, \frac{\beta}{2}}$, this implies
\begin{align}
\begin{split}
\adapt{p}(\noisymodelparami) =&\int_0^\infty \diff \varparami \adapt{p}(\varparami) \adapt{p}(\noisymodelparami \mid \varparami) \\
=& \int_0^\infty \diff \varparami 
\frac{1}{\Gamma \p{\frac{\beta}{2}}} 
\p{\frac{\beta}{2} \varparami e^{-\varparami}}^{\frac{\beta}{2}} \\
& \cdot \p{2 \pi \p{\varparami + \frac{1}{\beta}}}^{-\frac{1}{2}} e^{-\frac{1}{2 \p{\varparami + \frac{1}{\beta}}} \noisymodelparami^2}
\end{split}
\label{eq:beta-vae-probability}
\end{align}

Numerical results for the capacity $I(\noisymodelparami, (\modelparamsi))$ with varying $\beta$ are given below and plotted in \autoref{fig:beta-vae-capacity}.

\centering
\raisebox{-0.5\height}{\begin{tabular}{c c}
    $\beta$ & $I(\noisymodelparami, (\modelparamsi))$ \\
    \hline
    $0.01$ & $0.68\bits$ \\
    $0.1$ & $0.65\bits$ \\
    $1$ & $0.45\bits$ \\
    $10$ & $0.12\bits$ \\
    $100$ & $0.014\bits$
\end{tabular}}
    
    \newpage
    \bibliography{references}
    \bibliographystyle{icml2019}
\end{document}